# Conditional Expressions for Blind Deconvolution: Derivative form

S. Aogaki, I. Moritani, T. Sugai, F. Takeutchi, and F.M. Toyama, *Kyoto Sangyo University, Kyoto-603, Japan*

*Abstract*—We developed novel conditional expressions (CEs) for Lane and Bates' blind deconvolution. The CEs are given in term of the derivatives of the zero-values of the *z*-transform of given images. The CEs make it possible to automatically detect multiple blur convolved in the given images all at once without performing any analysis of the zero-sheets of the given images. We illustrate the multiple blur-detection by the CEs for a model image.

*Index Terms*-- Deconvolution, Image restoration.

## I. INTRODUCTION

Lane and Bates' (LB) blind deconvolution is based on the zero-sheets of the *z*-transform of given images [1]. It enables us to uniquely remove blurs convolved in the given images. However, due to its advanced analytical method the computational complexity of the image processing is large.

We developed novel and useful tool for the LB's blind deconvolution, i.e., conditional expressions (CEs) that make it possible to automatically find zero-values of blurs convolved in the given images. The CEs enable us to eliminate the blurs in the given images without performing any tough analysis of the zero-sheets of the images. We devised two versions of the CEs. One (version 1) is the CE that is given in terms of derivatives of zero-values evaluated at a single point. The other (version 2) is the CE that is given with zero-values evaluated at multi point. In this paper we present only the version 1 of the CEs and we relegate the version 2 to our second paper [2].

The version 1 of the CEs makes the LB's blind deconvolution perfectly automatic one. It can detect multiple blur all at once. In this paper we present the CE and illustrate the blur-detection by the CE for a model image.

In Sec. II we discuss the derivation of the CE. In Sec. III we illustrate the blur detection by the CE. Section IV is for the summary.

## II. CONDITIONAL EXPRESSIONS FOR MULTIPLE BLURS

We consider a situation where any noise is absent. In the situation an observed image $g(x,y)$ can be modeled as the convolution of a true image $f(x,y)$ and a blur image $h(x,y)$, i.e.,

$$g(x,y) = f(x,y) * h(x,y), \qquad (1)$$

where the blur $h(x,y)$ is assumed to be caused by a linear shift system. Our aim is to devise a mathematical tool for automatically finding the zero-values of the *z*-transform of $h(x,y)$. Throughout this paper, the sizes of the given image $g(x,y)$ and the blur $h(x,y)$ are denoted by $M \times N$ and $m \times n$, respectively. The *z*-transform $H(u,v)$ of $h(x,y)$ is written as

$$H(u,v) = \frac{1}{mn} \sum_{x=0}^{m-1} \sum_{y=0}^{n-1} h(x,y) u^x v^y, \qquad (2)$$

where $u$ and $v$ are complex variables. We first consider zero-values of $v$, which are the solutions of $H(u,v) = 0$ for a given $u$ and denoted by $\beta_i$ ($i = 1, 2, \cdots, n'; n' \le n-1$). Putting $u = \rho_u e^{-i\phi_u}$ ($\rho_u, \phi_u$: real parameters), $\beta_i$ are given as a function of $\rho_u$ and $\phi_u$, i.e., $\beta_i = \beta_i(\rho_u, \phi_u)$.

The zero-values $\beta_i$ satisfy the following $mn$ equations,

$$\begin{aligned} & H(\rho_u e^{-i\phi_u}, \beta_i) = 0, \\ & \frac{d}{d\xi_u} H(\rho_u e^{-i\phi_u}, \beta_i) = 0, \\ & \frac{d^2}{d\xi_u^2} H(\rho_u e^{-i\phi_u}, \beta_i) = 0, \\ & \vdots \\ & \frac{d^{mn-1}}{d\xi_u^{mn-1}} H(\rho_u e^{-i\phi_u}, \beta_i) = 0, \end{aligned} \qquad (3)$$

where $\xi_u$ stands for $\rho_u$ or $\phi_u$. Equation (3) is $mn$ simultaneous linear equations for $h(x,y)$. Coefficients of $h(x,y)$ in each equation of (3) are given in terms of $\xi_u$, $\beta_i$, $\partial \beta_i / \partial \xi_u$, $\partial^2 \beta_i / \partial \xi_u^2$, $\cdots$, and $\partial^{mn-1} \beta_i / \partial \xi_u^{mn-1}$. As the matrix $C$ composed of the coefficients of $h(x,y)$ is complex in general we relegate it to Appendix A, where we give the explicit form of $C$ for only $2 \times 3$ blurs, as a preparation to the discussion in the following.

We can construct a generator for the CE for $m \times n$ blurs in terms of the matrix $C$. The generator is given as $|C| = 0$, which is actually the condition for obtaining nontrivial solutions for $h(x,y)$ in (3). As the explicit forms of the CEs for blurs of large sizes are very complex, in the following we

S. Aogaki is with the Department of Information and Communication Sciences, (tel.: +81-75-705-1694, e-mail: aogaki@cc.kyoto-su.ac.jp).
I. Moritani is with the Department of Information and Communication Sciences, (tel.: +81-75-705-1694, e-mail: i654168@cc.kyoto-su.ac.jp).
T. Sugai is with the Department of Information and Communication Sciences, (tel.: +81-75-705-1694, e-mail: sugai@cc.kyoto-su.ac.jp).
F. Takeutchi is with the Department of Computer Sciences, (tel.: +81-75-705-1694, e-mail: takeut@ksuvx0.kyoto-su.ac.jp).
F.M. Toyama is with the Department of Information and Communication Sciences, (tel.: +81-75-705-1898, e-mail: toyama@cc.kyoto-su.ac.jp).

discuss with the CE for $2\times 3$ blurs, as a preparation to the next section.

From $|C|=0$ we obtain the following CE for $2\times 3$ blurs,

$$\begin{aligned}E^{\rho_u}_{2\times 3}(\beta_i) \equiv & -135\beta_i^{(2)^6} + 80\beta_i^{(1)^3}\beta_i^{(3)^3} + 15\beta_i^{(1)^4}\beta_i^{(4)^2} \\ & -60\beta_i^{(1)^3}\beta_i^{(2)}\beta_i^{(3)}\beta_i^{(4)} - 12\beta_i^{(1)^4}\beta_i^{(3)}\beta_i^{(5)} \\ & +18\beta_i^{(1)^3}\beta_i^{(2)^2}\beta_i^{(5)} + 270\beta_i^{(1)}\beta_i^{(2)^4}\beta_i^{(3)} \\ & -180\beta_i^{(1)^2}\beta_i^{(2)^2}\beta_i^{(3)^2} = 0,\end{aligned} \quad (4)$$

where $\beta_i^{(k)} \equiv \partial^k \beta_i / \partial \rho_u^k$. Equation (4) was obtained by taking $\xi_u = \rho_u$. If we take as $\xi_u = \phi_u$ we obtain another expression $E^{\phi_u}_{2\times 3}(\beta_i)$ of the CE for $2\times 3$ blurs. Although the form $E^{\phi_u}_{2\times 3}(\beta_i)$ so obtained is more complex than $E^{\rho_u}_{2\times 3}(\beta_i)$, they are in fact identical to each other in the sense that the two expressions are related as $E^{\phi_u}_{2\times 3}(\beta_i) = \rho_u^{12} E^{\rho_u}_{2\times 3}(\beta_i)$. This can be directly verified with the relations between derivatives of $\beta_i$ with respect to $\rho_u$ and $\phi_u$, $\partial \beta_i / \partial \phi_u = -i\rho_u (\partial \beta_i / \partial \rho_u)$ (the relations between higher degree derivatives are obtained from this relation). Therefore, in the following we discuss only the expression $E^{\rho_u}_{m\times n}(\beta_i)$. The CE $E^{\rho_u}_{2\times 3}(\beta_i)$ of (4) is evaluated for any single point of $\rho_u$ and $\phi_u$. In (3) we may use any higher degree derivatives of $H(u,v)$ with respect to $\xi_u$. However, it only makes the CE more complex. Equation (4) is the simplest version of the CE for $2\times 3$ blurs.

The $E^{\rho_u}_{2\times 3}(\beta_i)$ of (4) has been derived for $2\times 3$ blurs. However, $E^{\rho_u}_{2\times 3}(\beta_i)$ implicitly includes the CEs for blurs of the sizes smaller than $2\times 3$, i.e., $E^{\rho_u}_{2\times 2}(\beta_i)$, $E^{\rho_u}_{1\times 2}(\beta_i)$ and $E^{\rho_u}_{1\times 3}(\beta_i)$. This is because $E^{\rho_u}_{2\times 3}(\beta_i)$ can be expressed as a linear combination of $E^{\rho_u}_{2\times 2}(\beta_i)$ or $E^{\rho_u}_{1\times 3}(\beta_i)$. There exists no CE of $v$ for $i\times 1$ $(i=1,\cdots,M)$ blurs. (Note that there are many versions of the CE $E^{\rho_u}_{m\times n}(\beta_i)$ in general.) Further, $E^{\rho_u}_{2\times 2}(\beta_i)$ can be expressed as a linear combination of $E^{\rho_u}_{1\times 2}(\beta_i)$. Eventually, it is seen that $E^{\rho_u}_{2\times 3}(\beta_i)$ includes the CEs for $2\times 2$, $1\times 2$, and $1\times 3$ blurs implicitly. We relegate the explicit decomposition of $E^{\rho_u}_{2\times 3}(\beta_i)$ into $E^{\rho_u}_{2\times 2}(\beta_i)$ to [3]. Here, It is noted that there exists an additional case. The $E^{\rho_u}_{2\times 3}(\beta_i)$ also includes $E^{\rho_u}_{1\times k}(\beta_i)$ $(k=3+1,\cdots,N)$ for $1\times k$ blurs, which are partially larger than $2\times 3$. For such blurs, all zero-values $\beta_i$ $(i=1,\cdots,N'; N'\le N-1)$ are constants. Hence, any degree derivatives of $\beta_i$ are zero. This causes $E^{\rho_u}_{2\times 3}(\beta_i)=0$. This is why $E^{\rho_u}_{2\times 3}(\beta_i)$ includes $E^{\rho_u}_{1\times k}(\beta_i)$ implicitly. This feature of the CE holds for any $E^{\rho_u}_{m\times n}(\beta_i)$ in general. This is generally proven by basic manipulations for the determinant $|C|$. This feature is very useful because this enables us to automatically detect all zero-values of any blurs convolved in the given images, if we construct the CE for sufficiently large $m$ and $n$.

We have derived the CE by assuming the $m\times n$ blur $h(x,y)$. When we apply the CE to a given image, we need to evaluate derivatives of $\beta_i$ with respect to $\xi_u$ (i.e., $\rho_u$ or $\phi_u$) by using a given image $g(x,y)$ because we have no prior knowledge of the blur $h(x,y)$. This is possible as follows. For the zero-values of the blur $\beta_i$, $H(\rho_u e^{-i\phi_u},\beta_i)=0$. Hence, the derivatives $\partial \beta_i / \partial \xi_u$, $\partial^2 \beta_i / \partial \xi_u^2$, $\cdots$, and $\partial^{mn-1}\beta_i / \partial \xi_u^{mn-1}$ are respectively expressed as polynomial forms of $\beta_i$ from $dH(\rho_u e^{-i\phi_u},\beta_i)/d\xi_u = 0$, $d^2 H(\rho_u e^{-i\phi_u},\beta_i)/d\xi_u^2 = 0$, $\cdots$, and $d^{mn-1}H(\rho_u e^{-i\phi_u},\beta_i)/d\xi_u^{mn-1}=0$. On the other hand, the z-transform $G(u,v)$ of $g(x,y)$ is factorized as $G(u,v) = \sum_{x=0}^{M-1}\sum_{y=0}^{N-1} g(x,y)u^x v^y = F(u,v)H(u,v)$ where $F(u,v)$ is the z-transform of the true image $f(x,y)$. If $\beta_i$ is a zero-value of $H(\rho_u e^{-i\phi_u},\beta_i)$, then $H(\rho_u e^{-i\phi_u},\beta_i)=0$ and $F(\rho_u e^{-i\phi_u},\beta_i)\ne 0$. Accordingly, for $\beta_i$ of the blurs $d^k G(\rho_u e^{-i\phi_u},\beta_i)/d\xi_u^k = 0$ $(k=1,\cdots,mn-1)$ is respectively equivalent to $d^k H(\rho_u e^{-i\phi_u},\beta_i)/d\xi_u^k=0$. Hence, the analytical forms of $\partial^k \beta_i / \partial \xi_u^k$ $(k=1,\cdots,mn-1)$ needed for the CE are actually given in terms of $G(\rho_u e^{-i\phi_u},\beta_i)$ that is the z-transform of the given image. Thus, the derivatives of $\beta_i$ needed for the CE are all given as polynomials of $\beta_i$ themselves and the given image $g(x,y)$. We give some explicit forms of the derivatives in [3].

When we apply the CE to a given image, first we solve $G(\rho_u e^{-i\phi_u},v)=0$ for $v$ numerically to obtain $\beta_i$ $(i=1,2,\cdots,N'; N'\le N-1)$. Next we substitute them into the CE. As we mentioned above, the CE has been derived by using the z-transform $H(u,v)$ of $h(x,y)$. Nevertheless, we can evaluate the CE by using a given image $g(x,y)$. This fact makes the CE a very valuable tool. For any $\beta_i$ that are solved for the given image $g(x,y)$ numerically at any single point of $u$ we can automatically judge whether or not they are of the zero-values of the blurs without performing any tough analysis of the zero-sheets of the given image.

In order to restore a true image by eliminating blurs from the given image we need to repeat the same procedure also for zero-values $\gamma_i$ $(i=1,2,\cdots,M-1)$ of variable $u$. The CE for $\gamma_i$ are obtained in the similar manner by putting $v=\rho_v e^{-i\phi_v}$ in (2). For blurs of $m=n$, the form of the CE for $\gamma_i$ is the same as that for $\beta_i$. Therefore, once we obtain the CE for $\beta_i$ we can obtain the CE for $\gamma_i$ by the following replacements $\beta_i \to \gamma_i$, $\rho_u \to \rho_v$, $\phi_u \to \phi_v$, $x \rightleftarrows y$ and $m \rightleftarrows n$ in the CE for $\beta_i$ and relevant equations, but $g(x,y)$ must be left as it is. However, for blurs of $m\ne n$, the form of the CE for $\gamma_i$ is different from that of $\beta_i$. As a preparation to the next section, for the $2\times 3$ case we give the explicit form of the CE for $\gamma_i$, i.e.,

$$\begin{aligned}E^{\rho_v}_{2\times 3}(\gamma_i) \equiv & -40\gamma_i^{(3)^3} + 60\gamma_i^{(2)}\gamma_i^{(3)}\gamma_i^{(4)} - 15\gamma_i^{(1)}\gamma_i^{(4)^2} \\ & -18\gamma_i^{(2)^2}\gamma_i^{(5)} + 12\gamma_i^{(1)}\gamma_i^{(3)}\gamma_i^{(5)} = 0,\end{aligned} \quad (5)$$

where $\gamma_i^{(k)} \equiv \partial^k \gamma_i / \partial \rho_u^k$.

An advantage of the CEs is that for any zero-values numerically obtained at any single point of $u$ we can judge whether or not they are of blurs' ones by only substituting them into the CEs. Thus, the CEs are useful tools for an automatic blind deconvolution. On the other hand, the computational complexity becomes large as the size of the given image becomes large because many higher degree derivatives of the zero-values $\beta_i$ and $\gamma_i$ are needed to evaluate the CEs.

### III. ILLUSTRARTION

Fig.1 shows model images used for the illustration in this section. Fig. 1(a) shows a $40 \times 40$ model image that we regard as a true image, which has been often used in many references to demonstrate image restorations [4]. Figs. 1(b), 1(c), 1(d), and 1(e) represent blur images of the sizes $1 \times 2$, $2 \times 1$, $2 \times 2$, and $2 \times 3$, respectively. We convolved these four blurs into the true image of Fig. 1(a). Fig. 1(f) shows the convolved image, of which size is $43 \times 44$.

We illustrate how $E_{2\times3}^{\rho_u}(\beta_i)$ of (4) and $E_{2\times3}^{\rho_v}(\gamma_i)$ of (5) detect zero-values of the blurs. For the parameter $\rho_u$ and $\rho_v$ we took them to be $\rho_u = \rho_v = 1$. We can take any values for $\rho_u$ and $\rho_v$ in principle. However, in order to accomplish the image-restoration successfully we have to choose optimal values for $\rho_u$ and $\rho_v$. Fig. 2 shows the results of numerical evaluations of the CEs for the convolved image shown in Fig. 1(f). We carried out the evaluation at $\phi_u = 2\pi j / 43$ ($j = 0,\cdots,42$) and $\phi_v = 2\pi k / 44$ ($k = 0,\cdots,43$). Note that when we restore the image by the inverse Fourier transform we need the representation of $G(u,v)$ at those points of $\phi_u$ and $\phi_v$. In Fig. 2 we plotted $\log[|E_{2\times3}^{\rho_u}(\beta_i)|+1]$ and $\log[|E_{2\times3}^{\rho_v}(\gamma_i)|+1]$ for only four points of each $\phi_u$ and $\phi_v$. As we have stressed in the preceding section, both $E_{2\times3}^{\rho_u}(\beta_i)$ and $E_{2\times3}^{\rho_v}(\gamma_i)$ include CEs for blurs of smaller sizes $1\times2$, $2\times1$, $2\times2$, and $2\times3$ implicitly. Therefore, $E_{2\times3}^{\rho_u}(\beta_i)$ must detect totally four $(=1+1+2)$ zero-values $\beta_i$. When there exists a degenerate zero-value in the $2\times3$ blur, it is three $(=1+1+1)$. On the other hand, the number of zero-values $\gamma_i$ that should be detected by $E_{2\times3}^{\rho_v}(\gamma_i)$ is just three $(=1+1+1)$.

As seen in Fig. 2(a), for $\phi_u = 0$, four zero-values $\beta_1$, $\beta_{13}$, $\beta_{20}$, and $\beta_{21}$ are detected by $E_{2\times3}^{\rho_u}(\beta_i)$. Also at other $\phi_u$ four zero-values are well detected by $E_{2\times3}^{\rho_u}(\beta_i)$. Here, note that the solution-numbers of the detected zero-values are different at each $\phi_u$. In the evaluation of the CE we solved $\beta_i$ numerically by *Mathematica*, separately at each $\phi_u$. Accordingly, the order of the numerical solutions (zero-values) is at random at each $\phi_u$. For $\phi_v = 0$, as seen in

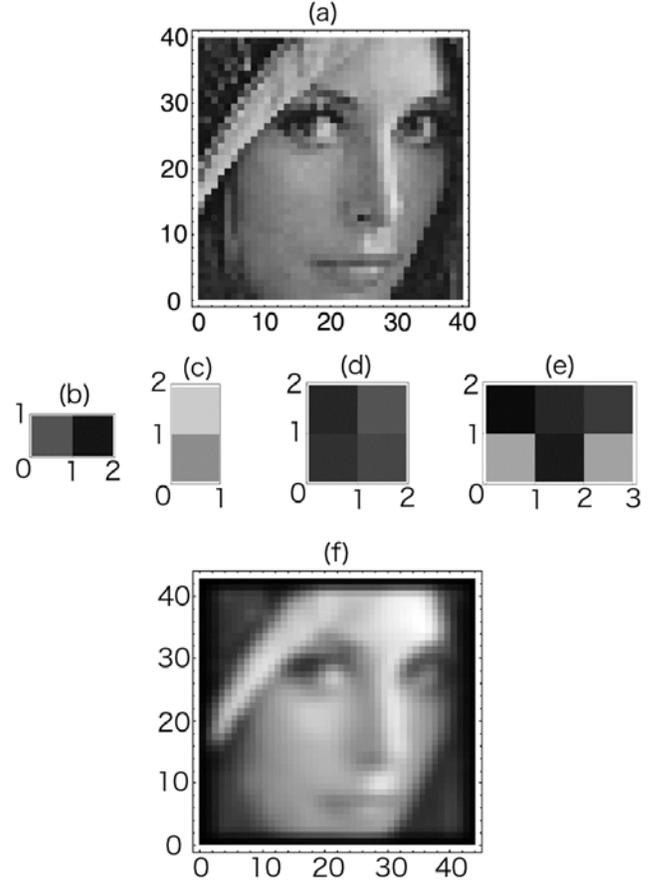

Fig. 1. (a): True image of $40 \times 40$ size that we took from [4]. (b): Blur image of $1\times 2$ size. (c): Blur image of $2\times 1$ size. (d): Blur image of $2\times 2$ size. (e): Blur image of $2\times 3$ size. (f): The image that was obtained by convolving the four blurs of (b)-(e) into the true image of (a). The size of the convolved image is $43 \times 44$.

Fig. 2(b), three zero-values $\gamma_1$, $\gamma_3$, and $\gamma_{12}$ are detected by $E_{2\times3}^{\rho_v}(\gamma_i)$, as expected. Also at other points $\phi_v$ three zero-values are well detected by $E_{2\times3}^{\rho_v}(\gamma_i)$. As we mentioned earlier, $E_{2\times3}^{\rho_u}(\beta_i)$ and $E_{2\times3}^{\rho_v}(\gamma_i)$ must detect the zero-values of blurs of not only the size $2\times 3$ but also the sizes $1\times 2$, $2\times 1$, and $2\times 2$, all at once. The results of Fig. 2 verify that the detection of the zero-values was done successfully.

In Fig. 3 we show the restored image by removing four zero-values $\beta_i$ in $v$ and three zero-values $\gamma_i$ in $u$. The restored image is perfectly the same as the true image of Fig. 1(a). Thus, we could confirm that $E_{2\times3}^{\rho_u}(\beta_i)$ and $E_{2\times3}^{\rho_v}(\gamma_i)$ certainly detect the zero-values of blurs convolved in the given image. The results show that the CEs would be very powerful mathematical tools for the LB's blind deconvolution.

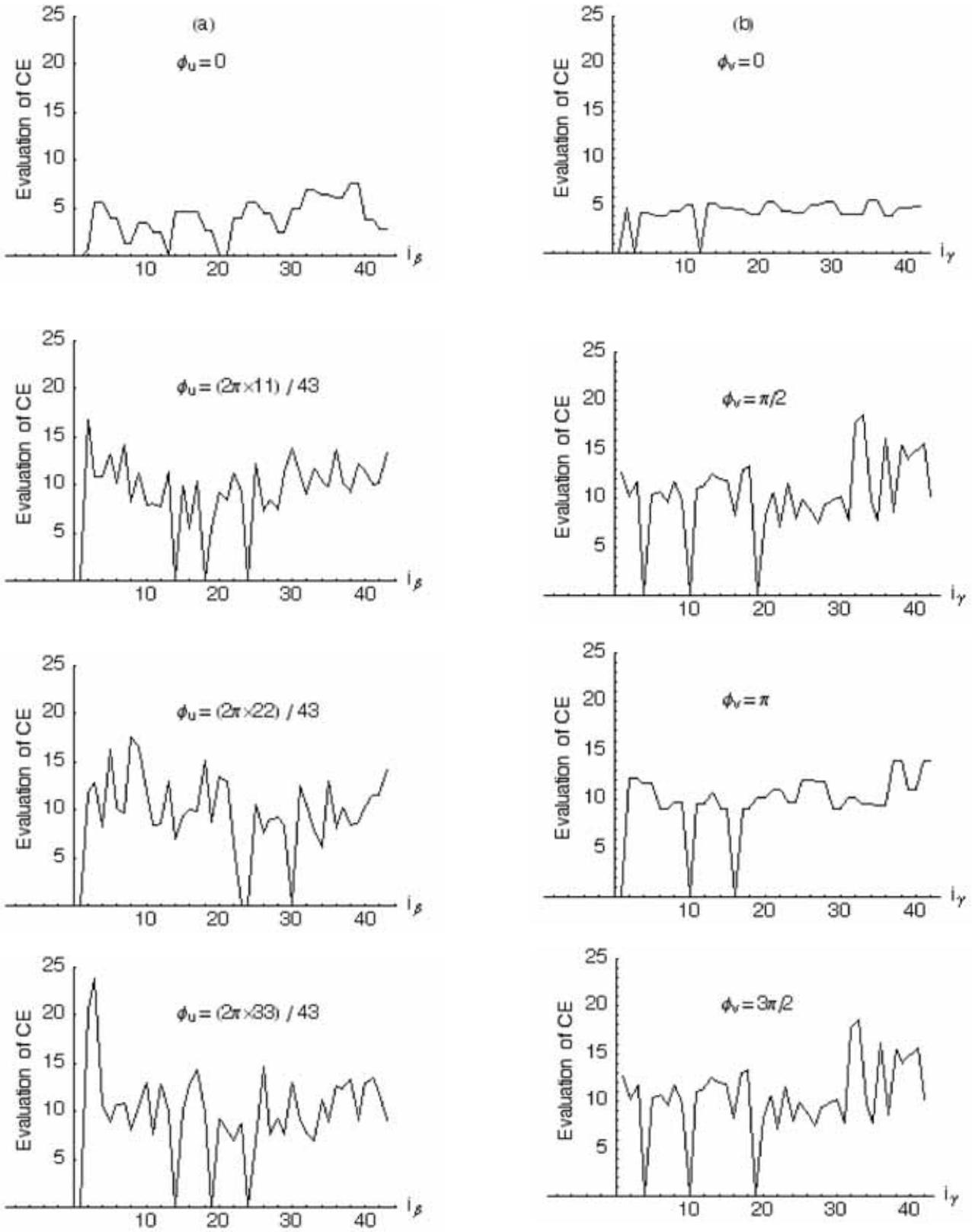

**Fig. 2.** Results of the evaluations of $E^{\rho_u}_{2\times 3}(\beta_i)$ and $E^{\rho_v}_{2\times 3}(\gamma_i)$ of (4) and (5). In (a) and (b), $\log[|E^{\rho_u}_{2\times 3}(\beta_i)|+1]$ and $\log[|E^{\rho_v}_{2\times 3}(\gamma_i)|+1]$ are plotted for the zero-value number $i_\beta$ and $i_\gamma$. We took $\rho_u$ and $\rho_v$ as $\rho_u = \rho_v = 1$.

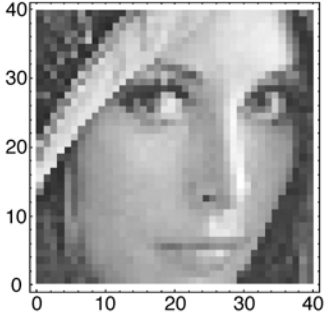

**Fig. 3.** Restored image by removing the four and three zero-values that were respectively detected by $E^{\rho_u}_{2\times 3}(\beta_i)$ and $E^{\rho_v}_{2\times 3}(\gamma_i)$.

IV. SUMMARY AND CONCLUTION

We have derived the CEs for finding blurs convolved in a given image, in a situation where any noise is absent. The CEs are given in terms of the derivatives of the zero-values of the *z*-transform of the given image. The derivatives of the zero-values are given as polynomials of the zero-values themselves. Hence, the CE can be evaluated with only the zero-values evaluated at any single point. This is a great advantage for the CEs. By using the CEs we can avoid a tough analysis of the zero-sheets of the given image and reduce the processing time.

The CEs constructed for $m \times n$ blurs can actually detect any blurs of the sizes smaller than $m \times n$ all at once. Therefore, when we apply the CEs to the given image, it is desirable to construct the CEs for blurs of sufficiently large size, although we have to take account of the computational complexity of the image restoration process. The CEs would be very useful to make the LB's blind deconvolution more practical one.

The CEs are very complex for blurs of large sizes. This causes a big computational load in the image restoration. In our second paper [2], we give the other version of the CEs that we devised to solve this problem. As we mentioned above the CEs detect multiple blur all at once. Instead the sizes of the detected blurs cannot be determined, except for a single blur. In our third paper [5] we present a simple method for finding only a single blur of a specified size $m \times n$.

**APPENDIX A: GENERATOR MATRIX** *C*

As an example, for a $2 \times 3$ blur we show the matrix $C$ composed of the coefficients of $h(x, y)$ in (3). The Matrix $C$ is given as

$$C = e^{2i\phi_u} \begin{pmatrix} 1 & \beta_i & \beta_i^2 & \rho_u & \rho_u \beta_i & \rho_u \beta_i^2 \\ 0 & \beta_i^{(1)} & 2\beta_i \beta_i^{(1)} & 1 & \beta_i + \rho_u \beta_i^{(1)} & \beta_i^2 + 2\rho_u \beta_i \beta_i^{(1)} \\ 0 & \beta_i^{(2)} & 2(\beta_i^{(1)^2} + \beta_i \beta_i^{(2)}) & 0 & 2\beta_i^{(1)} + \rho_u \beta_i^{(2)} & 2(2\beta_i \beta_i^{(1)} + \rho_u \beta_i^{(1)^2} + \rho_u \beta_i \beta_i^{(2)}) \\ 0 & \beta_i^{(3)} & 2(3\beta_i^{(1)} \beta_i^{(2)} + \beta_i \beta_i^{(3)}) & 0 & 3\beta_i^{(2)} + \rho_u \beta_i^{(3)} & 2(3\beta_i^{(1)^2} + 3\beta_i \beta_i^{(2)} + 3\rho_u \beta_i^{(1)} \beta_i^{(2)} + \rho_u \beta_i \beta_i^{(3)}) \\ 0 & \beta_i^{(4)} & 2(3\beta_i^{(2)^2} + 4\beta_i^{(1)} \beta_i^{(3)} + \beta_i \beta_i^{(4)}) & 0 & 4\beta_i^{(3)} + \rho_u \beta_i^{(4)} & 2(12\beta_i^{(1)} \beta_i^{(2)} + 4\beta_i \beta_i^{(3)} + 3\rho_u \beta_i^{(2)^2} + 4\rho_u \beta_i^{(1)} \beta_i^{(3)} + \rho_u \beta_i \beta_i^{(4)}) \\ 0 & \beta_i^{(5)} & 2(10\beta_i^{(2)} \beta_i^{(3)} + 5\beta_i^{(1)} \beta_i^{(4)} + \beta_i \beta_i^{(5)}) & 0 & 5\beta_i^{(4)} + \rho_u \beta_i^{(5)} & 2(15\beta_i^{(2)^2} + 20\beta_i^{(1)} \beta_i^{(3)} + 5\beta_i \beta_i^{(4)} + 10\rho_u \beta_i^{(2)} \beta_i^{(3)} + 5\rho_u \beta_i^{(1)} \beta_i^{(4)} + \rho_u \beta_i \beta_i^{(5)}) \end{pmatrix}, \quad (A1)$$

where we took as $\xi_u = \rho_u$ and $\beta_i^{(k)} \equiv \partial^k \beta_i / \partial \rho_u^k$. The CE $E^{\rho_u}_{2\times 3}(\beta_i)$ of (4) is obtained from $|C| = 0$.

ACKNOWLEDGMENT

This work was supported in part by the Science Research Promotion from the Promotion and Mutual Aid Corporation for private Schools of Japan, and a grant from Institute for Comprehensive Research, Kyoto Sangyo University.